\documentclass{article}
\usepackage{spconf,amsmath,amssymb,graphicx,booktabs}
\usepackage{xcolor}

\definecolor{batred}{RGB}{176,42,52}
\definecolor{batgray}{RGB}{242,242,242}
\newcommand{\batval}[1]{{\setlength{\fboxsep}{0.5pt}\colorbox{batgray}{\strut\textcolor{batred}{\textbf{#1}}}}}
\title{BEHAVIORAL AFFINITY TRANSDUCTION FOR ZERO-SHOT VISION-LANGUAGE CLASSIFICATION}
\name{
\shortstack{Junxi Huang$^{1}$, Wenshuo Chen$^{2}$, Xiucheng Wang$^{1}$, Meitong Shen$^{1}$,\\
Shaofeng Liang$^{2}$, Yuehan Kou$^{1}$, Yutao Yue$^{2,\dagger}$, Nan Cheng$^{1,*}$}\thanks{Corresponding authors are $^{*}$Nan Cheng (dr.nan.cheng@ieee.org) and $^{\dagger}$Yutao Yue (yutaoyue@hkustgz.edu.cn).}
}
\address{
{\small $^{1}$School of Telecommunications Engineering, Xidian University, Xi'an, Shaanxi 710071, China}\\
{\small $^{2}$The Hong Kong University of Science and Technology (Guangzhou), Guangzhou, China}
}

\begin{document}
\maketitle

\begin{abstract}
An image that is hard to classify on its own may have a helpful neighbor in the same unlabeled batch. Zero-shot vision-language models (VLMs) score images independently and leave this evidence unused. We study small unlabeled query batches with no labeled examples. Visual similarity can reflect pose or background, so it may connect images from different classes. Each image's scores over the class vocabulary offer another way to choose neighbors. We introduce Behavioral Affinity Transduction (BAT), which builds a sparse graph from these class-response profiles. The graph supplies neighbor evidence during prediction. For EM-Dirichlet, it also screens candidate classes and initializes assignments. BAT plugs into direct and iterative readouts of a frozen VLM. Across diverse datasets, BAT improves the mean accuracy of every tested readout. In the main RN50 comparison, soft and hard EM-Dirichlet gain $4.22$ and $2.72$ percentage points.
\end{abstract}


\section{Introduction}
\label{sec:introduction}

Consider a small album of pet photos. One beagle is partly hidden by a blanket, while another is clearly visible. A terrier in the same room may resemble the hidden beagle more because of the background. A zero-shot vision-language model (VLM) such as CLIP scores each photo against breed names independently \cite{radford2021clip}. The clear beagle could help label the hidden one if the model knew which photos to compare. Video clips and photo collections often provide such unlabeled groups. Transductive inference uses the group to refine predictions jointly.

\begin{figure}[t]
    \centering
    \includegraphics[width=\columnwidth]{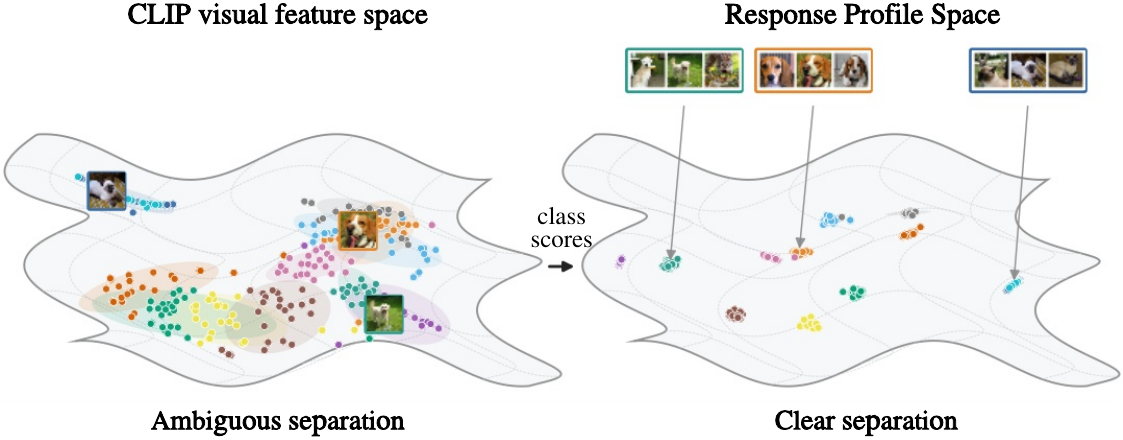}
    \caption{OxfordPets visual and response t-SNE projections.}
    \label{fig:motivation}
\end{figure}

We study one unlabeled batch of $N=75$ images with a frozen VLM and the full class vocabulary. There are no labeled support images or other test samples. EM-Dirichlet fits class distributions to batch responses \cite{martin2024transductiveclip}, and StatA adapts to the classes present \cite{zanella2025stata}. These methods estimate class usage across the batch. A difficult image still needs relevant peers. TransCLIP uses visual affinities \cite{zanella2024transclip}. ZLaP and ECALP propagate labels on image-text graphs \cite{stojnic2024zlap,li2025ecalp}, with ZLaP's transductive version using the full test set. In a batch of $75$, a misplaced edge can pass evidence from the wrong image.

Occlusion can separate the beagles in visual space. Their scores over the same breed names may still agree. Figure~\ref{fig:motivation} shows this pattern on OxfordPets. BAT uses the full class response to find images with similar plausible labels and joins them in a sparse graph.

\begin{figure}[t]
    \centering
    \includegraphics[width=\columnwidth]{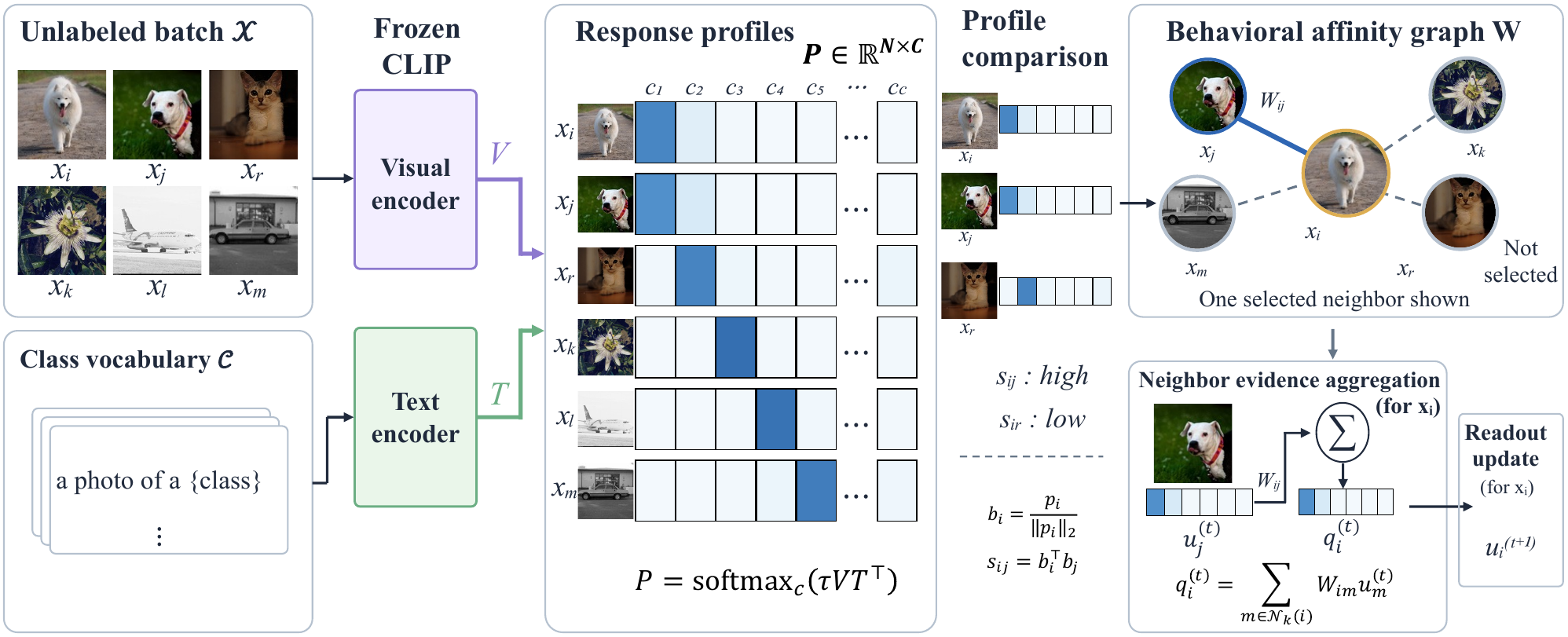}
    \caption{BAT graph construction and readout integration.}
    \label{fig:framework}
\end{figure}

BAT fits between the frozen VLM and a readout, as Fig.~\ref{fig:framework} shows. It builds one graph from cached class responses, then gathers neighboring assignments into a class message at each step. Direct prediction uses the message to revise scores. Iterative readouts use it to update assignments. EM-Dirichlet also uses the graph to screen candidate classes and initialize assignments. The encoders and class prototypes stay fixed, making BAT a plug-and-play module.

We make three contributions.
\begin{itemize}
\setlength{\itemsep}{0pt}
\setlength{\topsep}{2pt}
\item We build a sparse graph from class responses to link images with shared category evidence.
\item BAT shares one graph across seven readouts of a frozen VLM as a plug-and-play module.
\item At $N=75$, all seven readouts gain mean accuracy on ten datasets. Response graphs also outperform visual graphs for EM-Dirichlet.
\end{itemize}

\section{Behavioral Affinity Transduction}
\label{sec:method}

\subsection{Response Profiles}
\label{sec:response_profile}

Given $N$ unlabeled images and $C$ candidate classes, let $V\in\mathbb{R}^{N\times D}$ collect normalized visual features. Let $T\in\mathbb{R}^{C\times D}$ collect the corresponding text prototypes. The resulting class response matrix is
\begin{align}
P=\operatorname{softmax}_{c}\!\left(\tau VT^\top\right),
\label{eq:response_matrix}
\end{align}
where $\tau$ controls the response scale and softmax acts across classes. Each row $p_i=P_{i:}$ contains one image's scores over the vocabulary. We normalize it as $b_i=p_i/\lVert p_i\rVert_2$. The profile retains plausible alternatives beyond the top prediction.

\subsection{Behavioral Affinity Graph}
\label{sec:affinity_graph}

Two samples are neighbors when their response profiles agree across the vocabulary. Let $\mathcal{N}_k(i)$ contain the $\min(k,N-1)$ closest other query samples. BAT assigns their normalized affinity as
\begin{align}
W_{ij}=
\frac{
\mathbb{I}[j\in\mathcal{N}_k(i)]
\exp\!\left(b_i^\top b_j/\tau_g\right)
}{
\sum_{\ell\in\mathcal{N}_k(i)}
\exp\!\left(b_i^\top b_\ell/\tau_g\right)
},
\label{eq:affinity_graph}
\end{align}
where $\tau_g$ controls how strongly each row favors its closest neighbors. The sparse matrix satisfies $\sum_jW_{ij}=1$. Every profile uses the same class names, so each edge measures agreement in category evidence. BAT builds $W$ once and reuses it throughout inference.

The response overlap has a simple interpretation. If class labels are drawn independently from $p_i$ and $p_j$, their chance of agreement is $\sum_c p_{ic}p_{jc}$. BAT normalizes this overlap by the concentration of each response:
\begin{align}
b_i^\top b_j=
\frac{\sum_c p_{ic}p_{jc}}{\sqrt{\sum_c p_{ic}^2}\sqrt{\sum_c p_{jc}^2}}.
\label{eq:response_overlap}
\end{align}
Near one-hot responses connect matching classes and separate different classes. Ambiguous responses can agree through several plausible labels. Softmax also depends only on differences between class logits. Let $p(v)$ denote the class response of a normalized feature $v$. For any normalized $v'$ and scalar $a$,
\begin{align*}
(v'-v)T^\top=a\mathbf{1}_C
\quad\Longrightarrow\quad p(v')=p(v).
\end{align*}
Feature changes orthogonal to all differences between text prototypes satisfy this condition. Such changes can alter visual cosine similarity while leaving the response graph unchanged. BAT therefore selects neighbors in the geometry induced by the class vocabulary.

\begin{figure}[!b]
    \centering
    \includegraphics[width=\columnwidth]{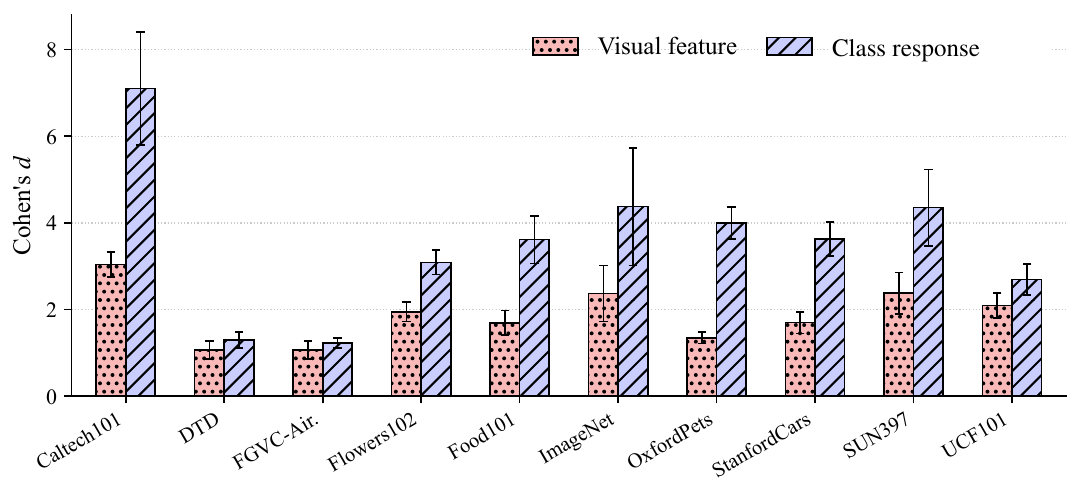}
    \caption{Visual and response class separation by Cohen's $d$.}
    \label{fig:cohen_d}
\end{figure}

\begin{figure*}[!t]
    \centering
    \includegraphics[width=\textwidth]{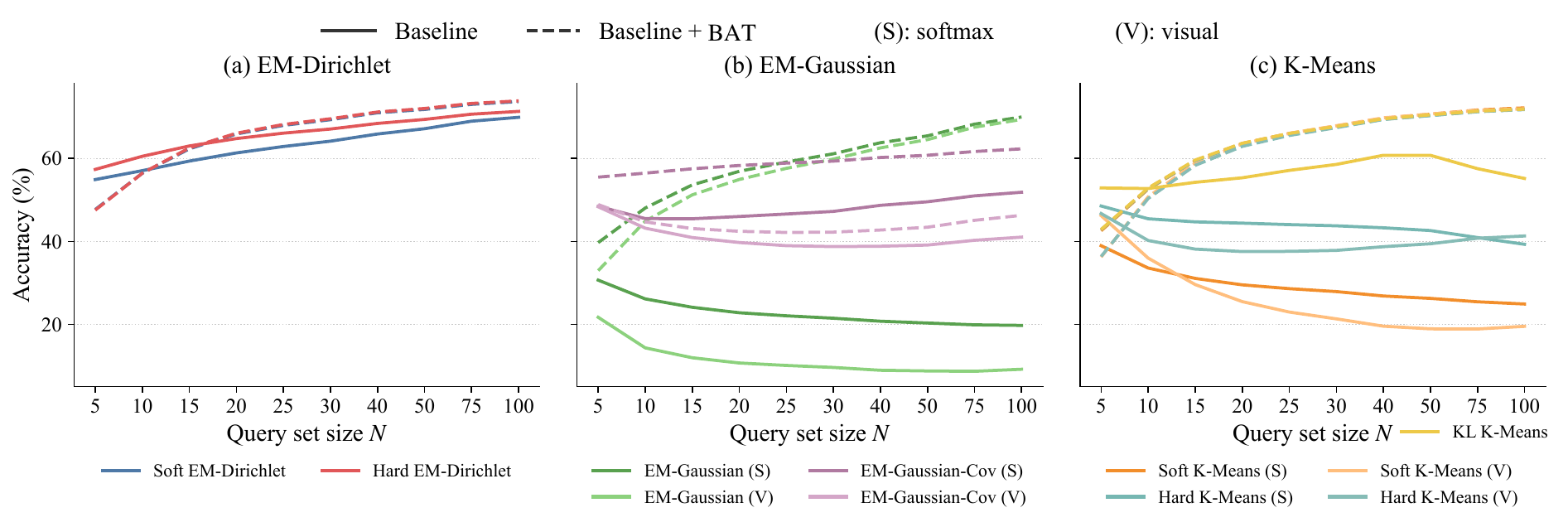}
    \caption{Accuracy across query sizes for EM-Dirichlet, Gaussian, and K-means readouts.}
    \label{fig:query_sensitivity}
\end{figure*}

\subsection{Integration Across Readouts}
\label{sec:readout_interface}

Let $U^{(t)}\in\mathbb{R}^{N\times C}$ denote the assignments at iteration $t$. Multiplication by $W$ gathers neighboring class evidence as $Q^{(t)}=WU^{(t)}$. For K-means and EM readouts, each method supplies its own scores $S^{(t)}$, and BAT adds graph evidence at the assignment step. Their updated scores and assignments are
\begin{align}
\widetilde S^{(t)}
&=S^{(t)}+\beta\log\!\left(Q^{(t)}+\epsilon\right),
\label{eq:generic_update}\\
U^{(t+1)}
&=\operatorname{softmax}_{c}\!\left(\widetilde S^{(t)}\right).
\label{eq:soft_assignment}
\end{align}
The coefficient $\beta$ controls graph evidence, and $\epsilon$ keeps the logarithm finite. For fixed neighbor messages, the softmax update maximizes $\langle u,S_{i:}^{(t)}+\beta\log(Q_{i:}^{(t)}+\epsilon)\rangle+H(u)$ over the class simplex. The entropy $H(u)=-\sum_c u_c\log u_c$ retains uncertainty. For two classes $c$ and $d$, the update gives
\begin{align}
\log\frac{u_{ic}^{(t+1)}}{u_{id}^{(t+1)}}
&=S_{ic}^{(t)}-S_{id}^{(t)}\nonumber\\
&\quad+\beta\log\frac{Q_{ic}^{(t)}+\epsilon}{Q_{id}^{(t)}+\epsilon}.
\label{eq:odds_update}
\end{align}
The readout score and neighbor support combine multiplicatively in the resulting assignment. A class overtakes another when its weighted neighbor log-odds exceeds the opposing score gap. Soft readouts keep the distribution. Hard readouts select its largest entry. K-means and Gaussian models apply the update during assignment. For direct prediction, BAT starts at $U^{(0)}=P$ and runs ten steps of $U^{(t+1)}=(1-\gamma)P+\gamma WU^{(t)}$ with $\gamma=0.5$. Because $W$ is row stochastic, this update converges to $U^\star=(1-\gamma)\sum_{r=0}^{\infty}\gamma^rW^rP$. Here $W^rP$ gathers evidence $r$ graph steps away, and $\gamma^r$ reduces the influence of longer paths.

\subsection{BAT with EM-Dirichlet}
\label{sec:em_dirichlet}

EM-Dirichlet fits a Dirichlet model to each class by alternating assignment and parameter updates \cite{martin2024transductiveclip}. For input $x_i=p_i$ and class parameters $\alpha_c^{(t)}$, its score is
\begin{align}
\ell_{ic}^{(t)}=\log\operatorname{Dir}\!\left(x_i;\alpha_c^{(t)}\right)
+\lambda v_c^{(t)}/N,
\label{eq:em_score}
\end{align}
where $\lambda=N\lfloor C/5\rfloor$ and $v_c^{(t)}$ is the log batch mean assignment to class $c$, shifted by one. This term captures overall class usage. BAT first screens candidate classes using neighbor predictions. In the initial assignments, classes absent from this set are set to $\epsilon$ before each row is normalized to obtain $\widetilde P$. Graph propagation then initializes EM through
\begin{align}
U^{(0)}&=\widetilde P, \nonumber\\
U^{(r+1)}&=\eta U^{(r)}+(1-\eta)WU^{(r)},
\quad r=0,\ldots,R-1,
\label{eq:graph_initialization}
\end{align}
where $\eta$ balances each response with its neighborhood. At each EM iteration, the graph adds neighboring assignments to the Dirichlet scores,
\begin{align}
u_{i:}^{(t+1)}=\operatorname{softmax}_{c}\!\left(
\ell_{i:}^{(t)}+\beta\log\!\left(Q_{i:}^{(t)}+\epsilon\right)
\right).
\label{eq:bat_em_update}
\end{align}
BAT computes $Q^{(t)}$ from the preceding assignments and updates all samples in parallel. Soft BAT keeps the complete distribution. Hard BAT selects one class per sample.

Response computation costs $\mathcal{O}(NDC)$, and pairwise affinity costs $\mathcal{O}(N^2C)$. The graph stores $\mathcal{O}(Nk)$ edges, so each update costs $\mathcal{O}(NkC)$. All stages operate on cached VLM responses with one shared graph.

\begin{table}[t]
\centering
\caption{EM-Dirichlet~\cite{martin2024transductiveclip} accuracy on RN50 (\%).}
\label{tab:main_results}
\begingroup
\setlength{\tabcolsep}{2.5pt}
\begin{tabular}{@{}lrrrr@{}}
\toprule
& \multicolumn{2}{c}{Soft} & \multicolumn{2}{c}{Hard} \\
\cmidrule(lr){2-3}\cmidrule(l){4-5}
Dataset & Base & BAT & Base & BAT \\
\midrule
Caltech101 & 88.42 & \batval{91.28} & 89.54 & \batval{91.29} \\
DTD & 47.17 & \batval{49.76} & 48.66 & \batval{50.27} \\
Aircraft & 19.25 & \batval{22.16} & 20.38 & \batval{22.09} \\
Flowers102 & 71.62 & \batval{75.54} & 73.17 & \batval{75.51} \\
Food101 & 88.21 & \batval{93.01} & 90.39 & \batval{93.10} \\
ImageNet & 76.67 & \batval{82.26} & 77.32 & \batval{82.59} \\
Pets & 87.53 & \batval{93.59} & 91.08 & \batval{93.55} \\
Cars & 65.72 & \batval{71.81} & 68.09 & \batval{72.43} \\
SUN397 & 77.31 & \batval{81.14} & 78.58 & \batval{81.52} \\
UCF101 & 68.05 & \batval{71.55} & 69.70 & \batval{71.70} \\
\midrule
Average & 68.99 & \batval{73.21} & 70.69 & \batval{73.41} \\
\bottomrule
\end{tabular}
\endgroup
\end{table}

\begin{table}[!t]
\centering
\caption{Readout accuracy (\%) across ten datasets. Gains use unrounded means.}
\label{tab:readout_generality}
\begingroup
\setlength{\tabcolsep}{3pt}
\begin{tabular}{@{}lrrr@{}}
\toprule
Readout & Base & +BAT & Gain \\
\midrule
Direct prediction~\cite{radford2021clip} & 60.75 & \batval{68.22} & \batval{+7.47} \\
Soft K-means~\cite{banerjee2005bregman} & 25.26 & \batval{71.82} & \batval{+46.56} \\
Hard K-means~\cite{lloyd1982kmeans} & 41.19 & \batval{71.42} & \batval{+30.23} \\
KL K-means~\cite{chiaroni2024simplex} & 57.94 & \batval{71.53} & \batval{+13.60} \\
EM-Gaussian~\cite{bishop2006prml} & 19.69 & \batval{68.46} & \batval{+48.77} \\
Soft EM-Dir.~\cite{martin2024transductiveclip} & 68.99 & \batval{73.06} & \batval{+4.07} \\
Hard EM-Dir.~\cite{martin2024transductiveclip} & 70.69 & \batval{73.28} & \batval{+2.59} \\
\bottomrule
\end{tabular}
\endgroup
\end{table}

\section{Experiments}
\label{sec:experiments}

\subsection{Experimental Setup}
\label{sec:setup}

Table~\ref{tab:main_results} covers ten datasets \cite{feifei2006caltech101,cimpoi2014dtd,maji2013aircraft,nilsback2008flowers,bossard2014food101,russakovsky2015imagenet,parkhi2012pets,krause2013cars,xiao2010sun,soomro2012ucf101}. Each task samples three to ten classes and $N=75$ queries, retaining the full vocabulary. The main RN50 table averages $1{,}000$ tasks per dataset over five seeds for both methods. Other RN50 readout and ablation baselines use five seeds and BAT three. The matched graph-space control uses five seeds per graph. ViT-L/14 baselines use seed 0 and BAT uses three or four seeds by dataset. Frozen RN50 uses $\tau=30$, $k=5$, $\tau_g=0.05$, $\beta=15$, $\eta=0.5$, and $R=2$.

\subsection{Main Results}
\label{sec:main_results}

Across ten datasets, BAT raises soft EM-Dirichlet from $68.99\%$ to $73.21\%$ and hard EM-Dirichlet from $70.69\%$ to $73.41\%$ (Table~\ref{tab:main_results}). Both improve on every dataset. The largest soft gains occur on Cars ($6.09$ points), Pets ($6.06$), and ImageNet ($5.59$). The gains include fine-grained datasets and the broader ImageNet task.

\subsection{Readouts and Backbones}
\label{sec:generality}

All seven readouts gain mean accuracy at $N=75$ (Table~\ref{tab:readout_generality}). K-means and Gaussian readouts benefit most as graph messages guide assignments. EM-Dirichlet gains $4.07$ and $2.59$ points. With ViT-L/14, soft and hard BAT reach $81.71\%$ and $81.83\%$ (Table~\ref{tab:backbone}).

\begin{table}[t]
\centering
\caption{EM-Dirichlet~\cite{martin2024transductiveclip} by backbone (\%).}
\label{tab:backbone}
\begingroup
\setlength{\tabcolsep}{3pt}
\begin{tabular}{@{}llrrr@{}}
\toprule
Backbone & Readout & Base & BAT & Gain \\
\midrule
RN50 & Soft & 68.99 & \batval{73.06} & \batval{+4.07} \\
 & Hard & 70.69 & \batval{73.28} & \batval{+2.59} \\
ViT-L/14 & Soft & 77.89 & \batval{81.71} & \batval{+3.83} \\
 & Hard & 79.16 & \batval{81.83} & \batval{+2.67} \\
\bottomrule
\end{tabular}
\endgroup
\end{table}

\subsection{Graph and Component Analysis}
\label{sec:mechanism}

Response profiles improve the neighborhoods BAT uses. Cohen's $d$ rises from $1.87$ for visual features to $3.54$ for class responses (Fig.~\ref{fig:cohen_d}). Same@5 rises from $11.29\%$ to $14.64\%$, and weighted Same@5 from $17.62\%$ to $29.70\%$ (Table~\ref{tab:graph_quality}). Oracle ratio normalizes Same@5 by its task-wise maximum.

A matched five-seed control at $N=75$ favors response graphs by about half a point for both EM-Dirichlet modes (Table~\ref{tab:graph_space}).

Removing refinement, initialization, or screening costs $1.88$, $0.52$, or $0.34$ points, respectively (Table~\ref{tab:ablation}). Refinement contributes at every assignment step.

\begin{table}[t]
\centering
\caption{Graph quality across ten datasets.}
\label{tab:graph_quality}
\begingroup
\setlength{\tabcolsep}{4pt}
\begin{tabular}{@{}lrr@{}}
\toprule
Metric & Visual & Response \\
\midrule
Same@5 (\%) & 11.29 & \batval{14.64} \\
Weighted Same@5 (\%) & 17.62 & \batval{29.70} \\
Oracle ratio (\%) & 67.26 & \batval{83.81} \\
\bottomrule
\end{tabular}
\endgroup
\end{table}

\begin{table}[!t]
\centering
\caption{EM-Dirichlet accuracy by graph space (\%).}
\label{tab:graph_space}
\begingroup
\setlength{\tabcolsep}{6pt}
\begin{tabular}{@{}lrr@{}}
\toprule
Graph space & Soft & Hard \\
\midrule
Visual & 71.62 & 71.89 \\
Response & \batval{72.13} & \batval{72.37} \\
\bottomrule
\end{tabular}
\endgroup
\end{table}

\begin{table}[!t]
\centering
\caption{Soft EM-Dirichlet component ablation (\%).}
\label{tab:ablation}
\begingroup
\setlength{\tabcolsep}{3pt}
\begin{tabular}{@{}lrr@{}}
\toprule
Configuration & Accuracy (\%) & Drop \\
\midrule
\batval{Full BAT} & \batval{73.06} & \batval{0.00} \\
No graph refinement & 71.18 & 1.88 \\
No graph initialization & 72.54 & 0.52 \\
No candidate screening & 72.72 & 0.34 \\
EM-Dirichlet~\cite{martin2024transductiveclip} baseline & 68.99 & 4.07 \\
\bottomrule
\end{tabular}
\endgroup
\end{table}

\begin{figure}[!t]
    \centering
    \includegraphics[width=\columnwidth]{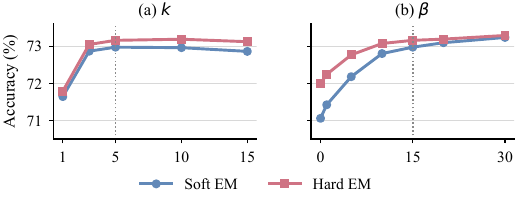}
    \caption{EM-Dirichlet accuracy sensitivity to $k$ and $\beta$.}
    \label{fig:sensitivity}
\end{figure}

\begin{table}[!ht]
\centering
\caption{EM-Dirichlet~\cite{martin2024transductiveclip} latency on an RTX 4080.}
\label{tab:latency}
\begingroup
\setlength{\tabcolsep}{4pt}
\begin{tabular}{@{}lrrr@{}}
\toprule
Readout & Base (ms) & BAT (ms) & Reduction (\%) \\
\midrule
Soft & 471.51 & 452.68 & 3.99 \\
Hard & 456.56 & 451.05 & 1.21 \\
\bottomrule
\end{tabular}
\endgroup
\end{table}

\subsection{Query Size, Sensitivity, and Cost}
\label{sec:robustness}

Figure~\ref{fig:query_sensitivity} shows gains from $N=20$ onward, reaching $3.79$ and $2.53$ points for soft and hard EM-Dirichlet at $N=100$. Figure~\ref{fig:sensitivity} shows gains leveling off after $k=5$, with $\beta$ most influential at low values. A few compatible neighbors can change a class ranking. Table~\ref{tab:latency} shows latency reductions of $3.99\%$ and $1.21\%$.

\section{Conclusion}
\label{sec:conclusion}

Each image's class response helps identify useful neighbors in an unlabeled batch. BAT turns their assignments into evidence for seven readouts of a frozen VLM. At $N=75$, the response graph selects more same-label neighbors than a visual graph, and every readout gains mean accuracy across ten datasets.

\paragraph*{Acknowledgments.} Generative AI was used to translate Chinese text into English.

\clearpage
\let\batbibliography\thebibliography
\renewcommand{\thebibliography}[1]{\batbibliography{#1}\setlength{\itemsep}{0pt}}
\bibliographystyle{IEEEbib}
\bibliography{aaai2027}

\end{document}